\documentclass[10pt,journal,compsoc]{IEEEtran}
\ifCLASSOPTIONcompsoc
  \usepackage[nocompress]{cite}
\else
  \usepackage{cite}
\fi
\ifCLASSINFOpdf
  \usepackage[pdftex]{graphicx}
\else
  \usepackage[dvips]{graphicx}
\fi
\usepackage{amsmath}
\usepackage{amssymb}
\usepackage{algorithmic}

\usepackage{array}

\ifCLASSOPTIONcompsoc
  \usepackage[caption=false,font=footnotesize,labelfont=sf,textfont=sf]{subfig}
\else
  \usepackage[caption=false,font=footnotesize]{subfig}
\fi
\usepackage{url}

\usepackage{multirow}
\usepackage{booktabs}
\usepackage{rotating}
\usepackage{enumitem}
\usepackage{makecell}
\usepackage{threeparttable}
\usepackage{hyperref} 
\hypersetup{
hidelinks,
colorlinks=False,
linkcolor=black,
citecolor=black
}

\begin{document}
%
\title{Diff-ID: An Explainable Identity Difference Quantification Framework for DeepFake Detection}

\author{Chuer~Yu, Xuhong~Zhang, Yuxuan~Duan, Senbo~Yan, Zonghui~Wang, Yang~Xiang, Shouling~Ji, Wenzhi~Chen

\IEEEcompsocitemizethanks{
\IEEEcompsocthanksitem C. Yu, X. Zhang, Y. Duan, S. Yan, W. Chen, Z. Wang and S. Ji are with the College of Computer Science and Technology, Zhejiang University, Hangzhou, Zhejiang 310027, China; 
S. Ji is also with the school of Electrical and Computer Engineering, Georgia Institute of Technology, Atlanta, Georgia, 30332, USA. \protect\\ 
Email: \{yuchuer, zhangxuhong, duanyuxuan, senboyan, chenwz, zhwang, sji\}@zju.edu.cn
\IEEEcompsocthanksitem Y. Xiang is with the School of Software and Electrical Engineering, Swinburne University of Technology, Hawthorn, VIC 3122, Australia.\protect\\
E-mail: yxiang@swin.edu.au }
}

%
%

\markboth{Journal of \LaTeX\ Class Files,~Vol.~14, No.~8, March~2023}%
{Shell \MakeLowercase{\textit{et al.}}: Bare Demo of IEEEtran.cls for Computer Society Journals}
%



\IEEEtitleabstractindextext{%
\begin{abstract}
Despite the fact that DeepFake forgery detection algorithms have achieved impressive performance on known manipulations, they often face disastrous performance degradation when generalized to an unseen manipulation. Some recent works show improvement in generalization but rely on features fragile to image distortions such as compression. 
To this end, we propose \textit{Diff-ID}, a concise and effective approach that explains and measures the identity loss induced by facial manipulations.
When testing on an image of a specific person, \textit{Diff-ID} utilizes an authentic image of that person as a reference and aligns them to the same identity-insensitive attribute feature space by applying a face-swapping generator. 
We then visualize the identity loss between the test and the reference image from the image differences of the aligned pairs, and design a custom metric to quantify the identity loss. The metric is then proved to be effective in distinguishing the forgery images from the real ones.
Extensive experiments show that our approach achieves high detection performance on DeepFake images and state-of-the-art generalization ability to unknown forgery methods, while also being robust to image distortions. 
\end{abstract}

}

\maketitle

\IEEEdisplaynontitleabstractindextext

%
\IEEEpeerreviewmaketitle

\IEEEraisesectionheading{\section{Introduction}\label{sec:introduction}}

%
%
%
%
\IEEEPARstart{R}{ecent} advancements in deep generative models, especially Generative Adversarial Networks (GANs), are making generating fake faces more convenient while recognizing them more challenging. Face swapping, which replaces someone's face with another, is the most popular among the DeepFake manipulations. 
Abuse of this technology may cause serious harm to individuals and our society, e.g., DeepFake is being used for defaming the personality of celebrities and spreading fake content \cite{romano2018jordan,spivak2018deepfakes,harris2018deepfakes,botha2020fake}. Therefore, it is imperative to develop intelligent face-swapping detection technologies.

During the past few years, many CNN-based classifiers \cite{afchar2018mesonet,roessler2019faceforensicspp, li2019exposing,nguyen2019capsule} have been proposed to train on known DeepFake images in a supervised way to detect fake faces. These classifiers work pretty well and achieve impressive results on the seen manipulations. However, they usually suffer significant performance degradation when generalized to a new DeepFake manipulation. For example, methods \cite{roessler2019faceforensicspp,li2019exposing,nguyen2019capsule} trained on Deepfakes, one dataset generated by the synthesis algorithm named Deepfakes, have reached around 95\% AUC scores within the dataset but degraded to about 65\% on CelebDF, another dataset generated using an improved DeepFake synthesis algorithm. 

Detecting DeepFake samples is tricky. On the one hand, there are many image synthesis and manipulation approaches, such as Deepfakes\cite{Deepfakes}, FaceSwap\cite{FaceSwap}, Face2Face\cite{thies2016face2face}, NeuralTexture\cite{thies2019deferred}, FSGAN\cite{nirkin2019fsgan}, and DF-VAE\cite{jiang2020deeperforensics}, producing forgery images of various quality and with diverse artifacts. On the other hand, these approaches have defined different mask blending templates, e.g., full face, center face, or only mouth part, causing fake artifacts of distinct shapes and sizes. In a nutshell, forgery traces that remained in DeepFake samples are very unpredictable, making it difficult for detection models trained on certain types of DeepFake samples to generalize to samples with different fake traces.

To tackle this problem, there have been intense studies to improve the generalization ability of the detection model, like data augmentations \cite{wang2020cnn}, domain adaptation \cite{schonfeld2019generalized,cozzolino2018forensictransfer}, and patch-based classification to model local patterns \cite{chai2020makes}. However, these image-level methods are still overfitted to the seen manipulations. A few studies that propose to mine the universal difference\cite{luo2021generalizing, schwarcz2021finding} between the forgery images and the real ones or simulate it by blending specially transformed real faces\cite{li2020face, zhao2021learning, shiohara2022detecting}.
These works have achieved a clear generalization improvement. However, they use prior knowledge of existing forgery methods or specific datasets to summarize a set of universal forgery traces. With the advancement of forgery manipulations, these seemingly ubiquitous traces may not be well applicable to detecting unknown forgery samples. 
For example, the recently proposed head-swapping approach HeadSwapper\cite{shu2022few} leaves no such blending traces in the central face.
In addition, image compression and other distortions can easily destroy these low-level forgery traces, which weakens the models' practicability.

Another line of research focuses on video-level detection. Since fake faces in video are replaced frame by frame, it can result in discontinuity of the successive frames. Some approaches detect the irregular jitter\cite{gu2021spatiotemporal, zheng2021exploring, sun2021improving} of fake videos on the time series. 
Recently, there are studies proving the superior generalization ability of the continuous high-level semantic information\cite{haliassos2021lips, haliassos2022leveraging}, i.e., Natural Talking. Unfortunately, they cannot detect single images, which are the main form of communication in social media.

\begin{figure}
    \centering
    \includegraphics[width=\linewidth]{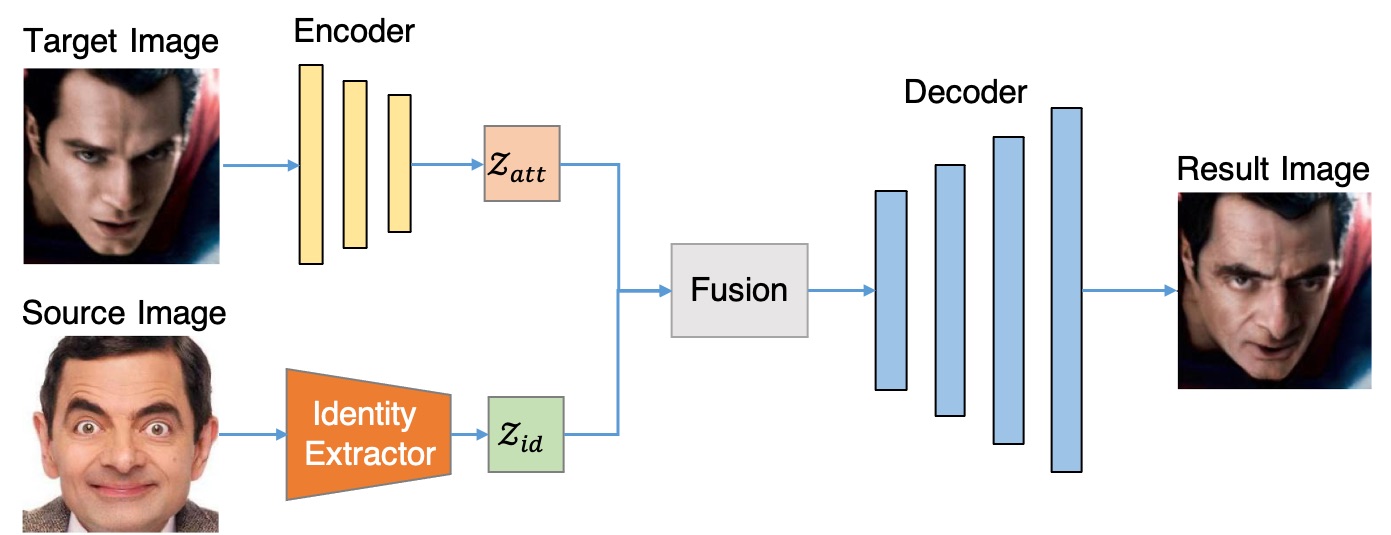}
    \caption{Face-swapping generator pipeline.}
    \label{fig:generator}
\end{figure}

Considering that the low-level traces of DeepFake images proposed so far are not competent in manipulation generalization and compression robustness at the same time, a natural question arises, ``is there a high-level semantic feature discriminating between real and fake images?'' Will it be robustly retained in the process of media propagation and show general forgery traces under different methods? {\bf We find that the identities of the fake images inevitably suffer from a non-negligible loss over the authentic identity.}
In the process of face swapping, shown in Figure~\ref{fig:generator}, the identity of the target image is replaced by the source, while the attributes of the target face, including head pose, expression, lighting, occlusions, hairstyle, and other background contents, are preserved\cite{li2021faceinpainter}.
\begin{figure}
    \centering
    \includegraphics[width=\linewidth]{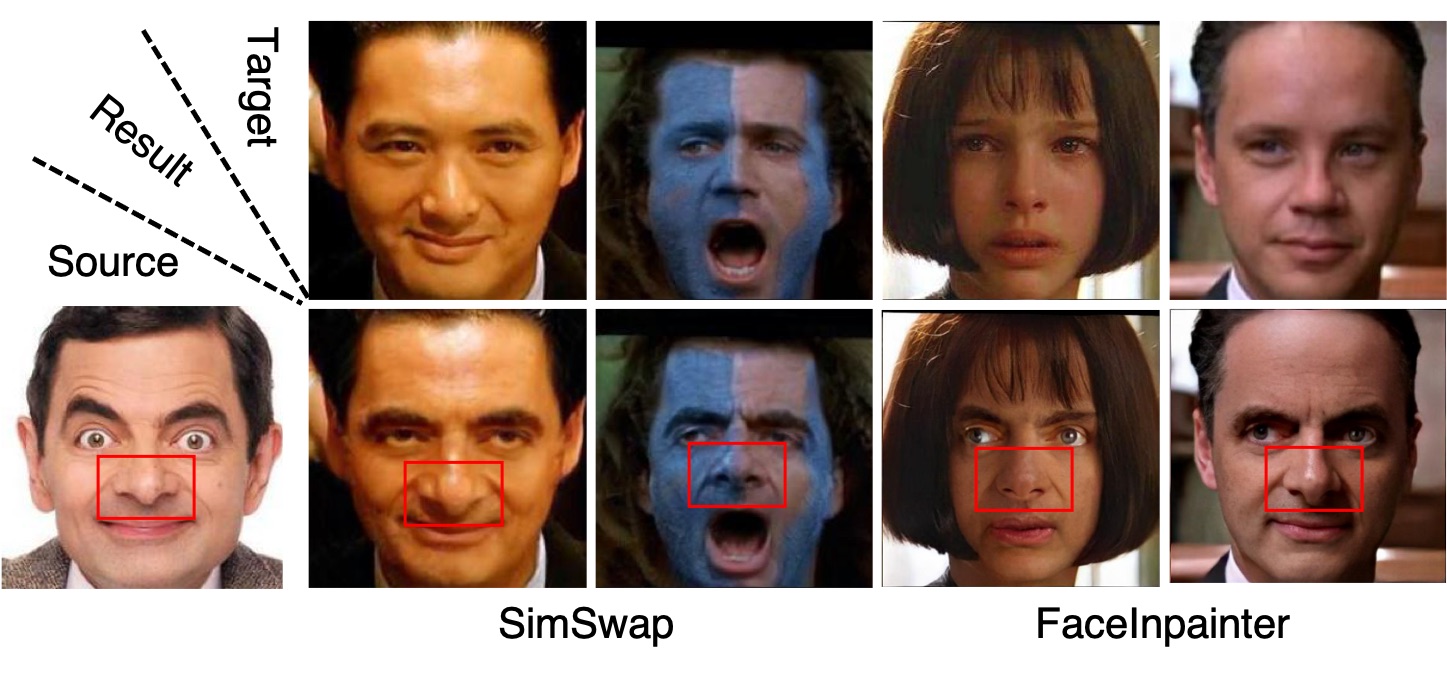}
    \caption{Face swapping examples generated by SimSwap \cite{chen2020simswap} and FaceInpainter \cite{li2021faceinpainter}. Red boxes outline the most significant changes in the identity feature. For illustration purposes, we select the most significant examples to reflect the identity loss. In real scenarios, DeepFake identity loss is hard to distinguish from the naked eye.}
    \label{fig:intro-diff-gan}
\end{figure}
However, we observe that identity features and some attribute features are deeply entangled. For example, the shape of eyebrows is an identity feature. It is simultaneously involved with attribute features like face shape and expression. 
Fusing the attribute features of the target may conflict with those entangled with the identity features of the source, causing identity loss of the source. 
Figure~\ref{fig:intro-diff-gan} shows forgery examples generated with the same source image using the latest methods \cite{chen2020simswap,li2021faceinpainter}. The red boxes outline the prominent identity feature, the nose in the example, which shows differences in the result forgery images. 
Due to the natural conflicts between the source identity features and the target attribute features, it is universal that the forgery image shows identity loss compared to the original identity features provided by the source image.

Intuitively, comparing the similarity of the identity embeddings of a test image and the reference image can infer whether the test image is forged. For example, we assume that there is a real reference face. If the identity embedding distance between the test face and the reference face is smaller than a set threshold, it is a real image; otherwise, it is a fake image. However, the extracted identity embedding is impure as it is deeply entangled with facial expressions, poses, age, and other attribute features\cite{chang2021learning, xu2022high}. We formulate the identity embedding of three face images of a particular person:
\begin{equation}
    \begin{aligned}
    \Phi_{id}&(I_{real1})=\mu^{id} + \varepsilon_1 \\
    \Phi_{id}&(I_{real2})=\mu^{id} + \varepsilon_2 \\
    \Phi_{id}&(I_{fake})=(\mu^{id} + \delta_{\mu^{id}}) + \varepsilon_3
    \end{aligned}
\end{equation}
where $\Phi_{id}$ is the face recognition model, $\mu^{id}$ denotes the ideal identity embedding of the person, $\delta_{\mu^{id}}$ denotes the identity loss in the fake face, and $\varepsilon_{i}$ denotes the additional attribute factors of each faces. It is possible that the additional attribute factors lead to greater discrepancy than the identity loss from fake faces. Since the attribute factors are variable and hard to estimate, detecting fake faces directly by identity embedding distance will bring a large error. Later in Section~\ref{section:4.3.1}, we will demonstrate this through experiments.

Therefore, we introduce \textit{Diff-ID}, which incorporates a reconstruction process that maps attribute factors $\varepsilon_{i}$ in identity embeddings to the same values $\varepsilon$ as in the reference or test image.
In this case, the identity differences are compared between $\Phi_{id}(I_{real}^{'})=\mu^{id} + \varepsilon$ and $\Phi_{id}(I_{fake}^{'})=\mu^{id} + \delta_{\mu^{id}} + \varepsilon$.
Thus, the identity loss $\delta_{\mu^{id}}$ caused by face-swapping manipulations could be well quantified.
Specifically, we provide \textit{Diff-ID} with an authentic reference image owning the same identity as the one under the test without constraints of special attribute features.
Obtaining such reference images is practical in the real world, as a forgery image usually has a claimed identity whose genuine photo is commonly available. For example, on LinkedIn\cite{LinkdIn}, every user has a registered photo, making protecting LinkedIn users practical. Some online platforms, e.g., Weibo\cite{weibo}, require users to authenticate their accounts with their national IDs, making referencing their ID photos possible. Besides, some social platforms such as Tinder\cite{Tinder}, Bumble \cite{Bumble}, and Badoo\cite{Badoo} (the most popular dating apps in Google Play), require registered users to take a selfie that matches a random pose on the screen for authentication (eliminate fake profile registrations). Correspondingly, these platforms can assume the responsibility of detecting whether there are DeepFake pictures of registered users spreading on the platform. Then, by applying a face-swapping generator, \textit{Diff-ID} aligns the test and reference images to the same identity-insensitive attribute space, including background, expression, posture, etc. The identity losses are well disclosed and visualized in pixel-level differences of the aligned generation results. Besides, the noise unrelated to the subtle identity loss is minimized by applying an adaptive fine-tuning scheme and a face-dedicated mask. Lastly, we designed a customized metric for identity loss to distinguish the forgery images from the real ones. 

Overall, our main contributions are summarized as follows.
\begin{itemize}
\item We reveal that the identities of fake faces suffer a non-negligible loss compared to real identities, which is then used as an explainable feature for detecting fake images.
\item We propose a method that visualizes and quantifies the identity difference between the forgery and authentic images of the same person for face-swapping detection.
\item Extensive evaluations indicate that our method has good generalization performance to unseen forgery samples and is robust to image distortions.
\item Our method is easy to deploy and resource-friendly. It requires only simple fine-tuning of the pre-trained generator and does not rely on fake samples or a large number of samples for training.
\end{itemize}

\section{Literature Review}
In this section, we discuss the related work that constitutes our present work's foundations and motivation.
\subsection{DeepFake Generation}
Common DeepFake generation techniques include entire image synthesis, modification of facial attributes, face identity swap, video puppeteering, etc. Face identity swap is one of the most popular used in recent years. It aims to integrate the identity of a source face into a target face while preserving the attribute of the target face, including head pose, expression, and other background contents.

The early approaches can only swap faces with similar postures \cite{bitouk2008face,wang2008facial}. After that, some 3D-based \cite{nirkin2018face,nirkin2019fsgan, natsume2018fsnet, cheng20093d} methods were proposed to fit a 3D morphable face model (3DMM) \cite{blanz1999morphable} and apply the expression components of one face to the other. In particular, Nirkin et al. \cite{nirkin2018face,nirkin2019fsgan} proposed a superior method using a fixed 3D face shape as the proxy and an occlusion-aware face segmentation network for face swapping. 

Recently, GAN-based \cite{Li2020faseshifter, chen2020simswap, zhu2021one, natsume2018rsgan, Deepfakes,korshunova2017fast,li2021faceinpainter,ijcai2021-157} approaches have shown great ability to generate vivid fake images. Early works like Deepfakes \cite{Deepfakes} and Korshunova et al. \cite{korshunova2017fast} model different source identities separately. However, these subject-specific methods are time-consuming as they require training specific decoders. 
To address this limitation, a line of subject-agnostic techniques, such as FaceShifter \cite{Li2020faseshifter}, SimSwap \cite{chen2020simswap}, FaceInpainter \cite{li2021faceinpainter}, and HifiFace \cite{ijcai2021-157}, are proposed to fuse the attribute of a target face and the identity of a source face in latent features for arbitrary face swapping. Furthermore, MegaFS \cite{zhu2021one} and HiRes \cite{xu2022high} use the pre-trained StyleGAN as a decoder to improve the forgery result to the Megapixel level.

With the advancement of DeepFake generation technology, the visual quality of the generated fake faces is improving, and the artifacts of blending two faces are almost undetectable. However, the phenomenon of identity and attribute entanglement on facial characteristics exists objectively. Therefore, identity loss exists in these forgery samples.

\subsection{DeepFake Detection}
DeepFake has attracted much attention over the past few years, with many works devoted to DeepFake detection. Early works capture visual clues like warping artifacts \cite{li2019exposing, matern2019exploiting}, inconsistent head poses \cite{yang2019exposing}, and color disparities \cite{li2020identification} of fake images. However, these visual flaws can be easily fixed as generation techniques improve. Then, some works are proposed to detect more indiscernible clues like digital GAN fingerprints \cite{nataraj2019detecting, yu2019attributing} or spectral distortions \cite{luo2021generalizing, liu2021spatial, qian2020thinking, li2021frequency}. Yet these works are easily overfitted on training data and perform unsatisfactorily on new forgery examples. 
Therefore, recent research has focused on finding more generalized clues for forgery detection. To tackle this, some methods propose using pristine images to reproduce common forgery artifacts \cite{li2020face, zhao2021learning, shiohara2022detecting}, like blending boundaries and statistical inconsistencies, whereas they are too fragile to image degradations.
There are also studies focusing on fake video detection, which explore the dynamic behavior between consecutive frames, e.g., face geometry consistency \cite{tursman2020towards, sun2021improving, gu2022delving}, heartbeat \cite{ciftci2020fakecatcher, qi2020deeprhythm, liang2021identifying}, lip movement \cite{yang2020preventing, haliassos2022leveraging, haliassos2021lips}. These methods have achieved noticeable improvements in the generalization of detection, but they cannot be applied to detect image frames. Our approach is a great addition to the image detection scene.

Another notable direction to improve generalization is verifying whether the behaviors/features of the videos/images to be tested are consistent with a given set of example real videos/images. We call this kind of study the \textbf{reference-assisted} forgery detection approach, where the reference specifically refers to the real, additional input with the same identity as the test sample. Our method is one such. The most related works are DISC\cite{jiang2021practical}, ID-Reveal\cite{cozzolino2021id}, and ICT\cite{dong2022protecting}. Specifically, DISC\cite{jiang2021practical} uses the spatial correlation within the query and reference images to generate an identity attention map and then digs deeper into these identity-related areas to extract forgery clues. ID-Reveal\cite{cozzolino2021id} estimates a temporal embedding of biometrical characteristics as a distance metric to distinguish fake videos. ICT\cite{dong2022protecting} calculates the distance between a suspect face's inner or outer identity vector and a reference to detect forgery. Such reference-assisted methods usually only protect specific groups of people (e.g., celebrities) and require relevant data of them for training. On the contrary, Our proposed method can be adapted to detect fake images of ordinary people, requiring only one real reference image of them. At the same time, our model does not depend on the images of the test persons for training. Therefore, it can generalize well on the open-set scenarios.

\section{Proposed Method}
In this section, we first formulate the insight of the design and then give a brief overview of our method, \textit{Diff-ID}. As aforementioned, DeepFake generators adaptively integrate the source person's identity into the target's attributes. We find out that \textit{some features entangled in the source identity may conflict with the target attributes and cause an identity loss in the generated fake image}.
Inspired by this, we explore and exploit the minor difference between the identity characteristics of the specific person in a test image and an authentic image, i.e., reference.

\begin{figure}
    \centering
    \includegraphics[width=\linewidth]{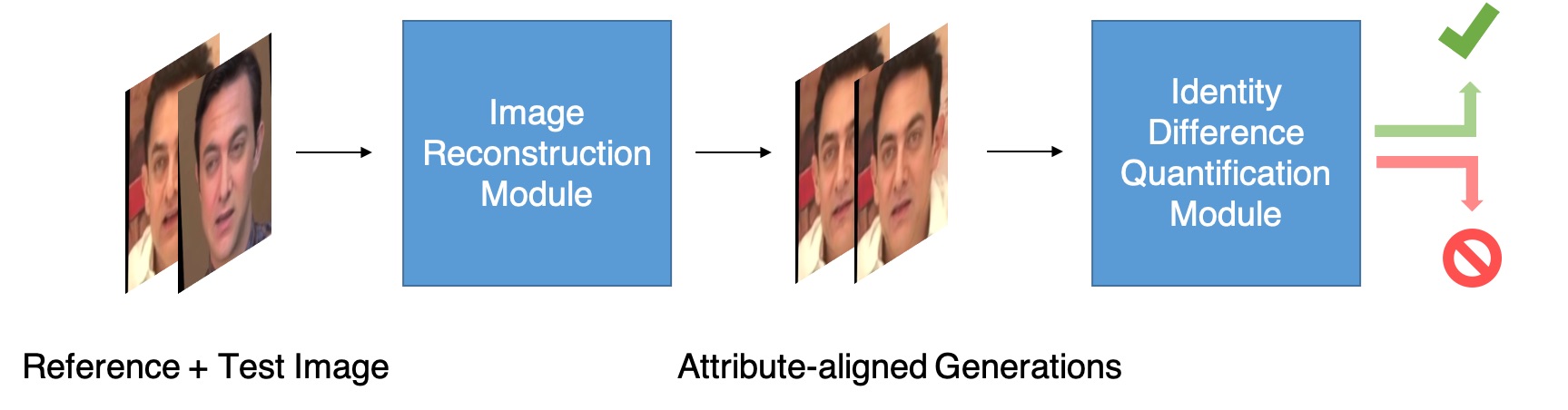}
    \caption{Overview of \textit{Diff-ID}. First, the reference and test image are sent to the image reconstruction module, where attribute-aligned generation results are obtained. Then the generations are sent to the Identity Difference Quantification Module to calculate a customized \textit{Diff-ID} metric that distinguishes real and fake samples.}
    \label{fig:framework}
\end{figure}

As depicted in Figure \ref{fig:framework}, \textit{Diff-ID} consists of two modules: an image reconstruction module and an identity difference quantification module.
When a suspected image is under test, an authentic image of the person it claims to be is selected as the reference. Then these two images are sent to the image reconstruction module to obtain attribute-aligned generations, which are all the possible face-swapping results of the identity feature and attributes features from both the test and reference images. After that, the identity difference quantification module calculates the identity loss by a customized metric, distinguishing a fake image from the real one.

In the following, we will describe the two modules in detail in Section~\ref{section:3.1} and Section~\ref{section:3.2}. Section~\ref{section:3.3} presents how to fine-tune an off-the-shelf face-swapping generator used in the image reconstruction module to fit our framework better.

\begin{figure}
    \centering
    \includegraphics[width=\linewidth]{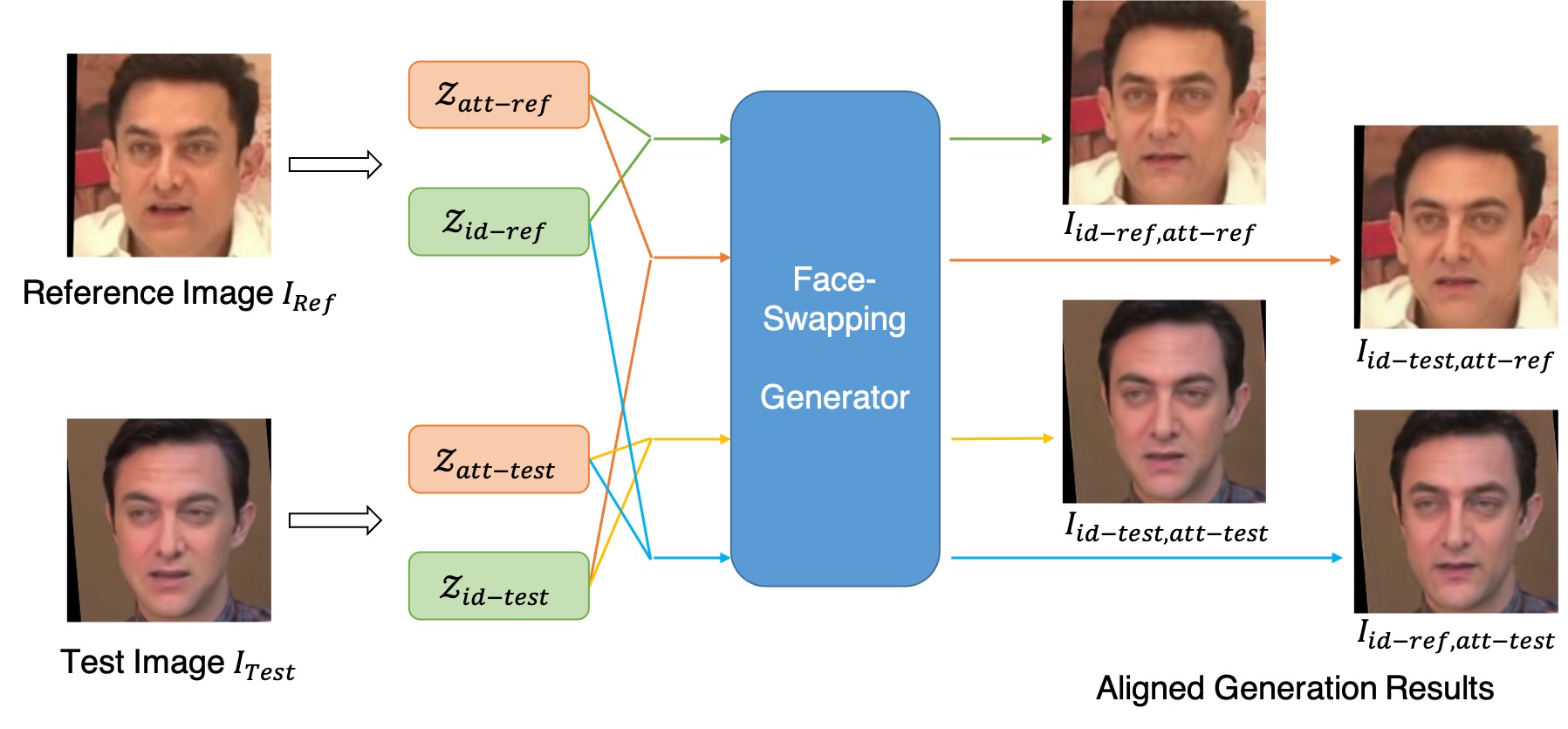}
    \caption{The Image Reconstruction Module. Through reconstruction, the reference and test images can be aligned to either the attribute space of the reference image or that of the test image.}
    \label{fig:img_recon}
\end{figure}
\subsection{Image Reconstruction Module}
\label{section:3.1}
Figure \ref{fig:img_recon} displays the image reconstruction module. It aligns the test image and the reference to the same attribute space, including background, facial expression, posture, etc. The core function is provided by a face-swapping generator, which is represented as $G$ below. It extracts identity embedding from the source image and attribute embedding from the target image and then makes an adaptive fusion of them to generate the face-swapping result.

For a test image $I_{Test}$ claiming to be a specific person, taking Aamir Khan, an actor from Indian Bollywood, as an example, we choose an authentic image of him as the reference $I_{Ref}$. 
Following the process depicted in Figure~\ref{fig:img_recon}, we extract the attribute embedding $\mathcal{Z}_{att}$ and identity embedding $\mathcal{Z}_{id}$ from both $I_{Ref}$ and $I_{Test}$. We use $\mathcal{Z}_{id-ref}$ to represent the identity embedding extracted from the reference image $I_{Ref}$. Similarly, $\mathcal{Z}_{att-test}$ denotes the attribute embedding extracted from the test image $I_{Test}$. Then, we select one identity and one attribute embedding and feed the embedding pairs (e.g., $(\mathcal{Z}_{id-ref}, \mathcal{Z}_{att-test})$) into the generator $G$. As there are one reference and one test image, we can extract two identity embeddings and two attribute embeddings in all, resulting in four different embedding pair combinations and, thus, four generated images. We denote them by:
\begin{equation}
    \begin{aligned}
    I&_{id-ref,att-ref}=G(\mathcal{Z}_{id-ref}, \mathcal{Z}_{att-ref}) \\
    I&_{id-test,att-ref}=G(\mathcal{Z}_{id-test}, \mathcal{Z}_{att-ref}) \\
    I&_{id-test,att-test}=G(\mathcal{Z}_{id-test}, \mathcal{Z}_{att-test}) \\ 
    I&_{id-ref,att-test}=G(\mathcal{Z}_{id-ref}, \mathcal{Z}_{att-test}) \\
    \end{aligned}
\end{equation}

As an example, $I_{id-ref,att-test}$ represents the generation result of the identity embedding provided by $I_{Ref}$ and the attribute embedding provided by $I_{Test}$.
Among the generations, $I_{id-ref,att-ref}$ and $I_{id-test,att-ref}$ share the same attribute feature as $I_{Ref}$. However, the two images show slight differences in facial details, as they are from two slightly different identity embedding inputs. Given that the reconstruction process is symmetrical, similar observations can be obtained among $I_{Test}$, $I_{id-test,att-test}$, and $I_{id-ref,att-test}$. The following section introduces how to evaluate the identity difference using the interrelation between these images.

\subsection{Identity Difference Quantification Module}
\label{section:3.2}
The identity difference quantification module assesses how large the identity features of the test and reference images differ.
\begin{figure}
    \centering
    \includegraphics[width=\linewidth]{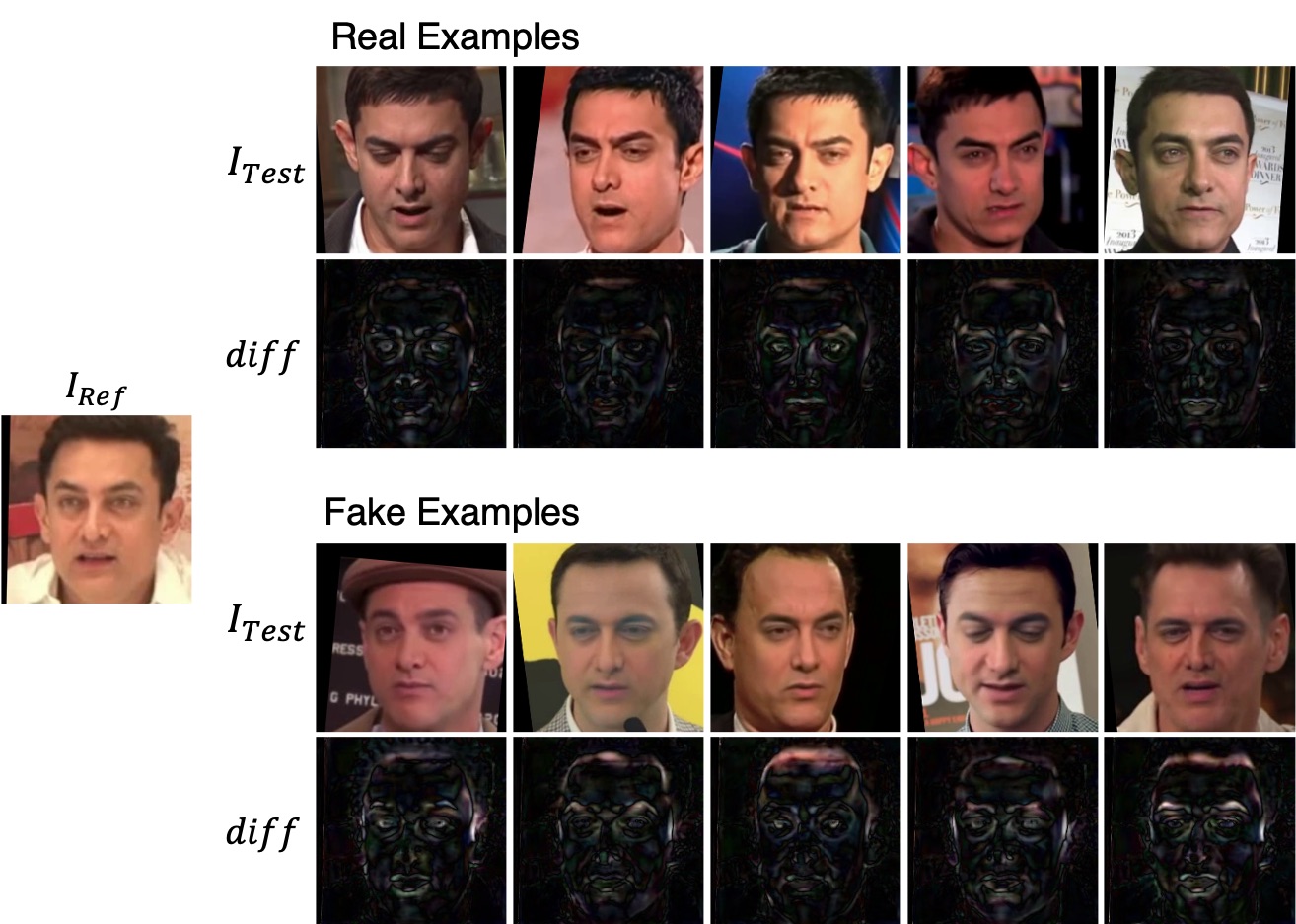}
    \caption{Examples of $diff(I_{id-ref,att-ref},I_{id-test,att-ref})$ when the test images are real/fake. The diff images are magnified five times for better visual display.}
    \label{fig:examples}
\end{figure}

\textbf{Empirical study}. 
As clarified before, $I_{id-ref,att-ref}$ (a self-reconstruction of the reference image) and $I_{id-test,att-ref}$ share the same attribute while showing differences in identity in terms of facial details. We consider calculating the pixel differences between the two generations. 
We formulate the pixel differences of two images $I_1$ and $I_2$ as follows:
\begin{align}
    diff(I_1, I_2)=|I_1-I_2|
\end{align}
We also refer to the image pixel differences as $diff$ for short.
In this way, we formulate the image differences of the two generated images as $diff(I_{id-ref,att-ref}, I_{id-test,att-ref})$, which reveals the identity differences under the same attribute space from the reference image. 
In the following, we use some example test images to visualize the calculated $diff$.
In Figure~\ref{fig:examples}, take Aamir Khan as an example, we select one real image of him as the reference and some real and fake images of him as the test images. Then we generate $I_{id-test,att-ref}$ following the process in the image reconstruction module and display the five times pixel differences $diff(I_{id-ref,att-ref}, I_{id-test,att-ref})$ (labeled $diff$ in the figure) below the example test images. 
A large value, i.e., a bright pixel in the $diff$ image, indicates a noticeable gap between the identities of reference and test images. Compared to the real images, we notice that the fake images show more evident differences with the reference image in sideburns, eyebrows, forehead, or other identity characteristics. This aligns with our intuition: \textit{for a specific person, the difference in identity features of real faces is usually smaller than that between real and fake}. 
As a result, a simple accumulation of pixel-level differences (i.e., L2 norm) can empirically classify real and fake images.

\textbf{Formulation}.
Based on the empirical study, the pixel differences of images visualize the face region where the identity features have differences, as we have aligned images to the same attribute space. The accumulation of it explains how large the identity features of the reference and the test image differ.
However, an isolated difference may not represent the loss of identity characteristics well. 
Therefore, we consider a normalized identity loss.
We find that the image reconstruction loss positively correlates with and reflects the richness and granularity of the identity feature. Therefore, we divide the identity loss by the image reconstruction loss to evaluate the normalized identity loss. 
In the meanwhile, we also observed that the generator has difficulty reconstructing complex textures, which would introduce much noise in accumulating pixel differences. Hence, we deliver a mask matrix to filter out the noise outside the face region and focus on capturing identity differences.
In this way, we formulate a set of image differences as pixel-level L2 distances:
\begin{equation}
    \begin{aligned}
    l&_{ref:recon} = ||M_{ref} \odot (I_{id-ref,att-ref} - I_{Ref})||_2 \\
    l&_{ref:recon+id} = ||M_{ref} \odot (I_{id-test,att-ref} - I_{Ref})||_2 \\
    l&_{ref:id} = ||M_{ref} \odot (I_{id-ref,att-ref} - I_{id-test,att-ref})||_2 \\
    \end{aligned}
\end{equation}
Where $\odot$ means dot product, $M_{ref}$ is the binary mask of image $I_{Ref}$ in which value 1 denotes the face region and value 0 denotes the outer identity-independent background. To get the mask, we apply the face parsing tool~\cite{face-parsing} to detect the face area and use the Gaussian kernel to dilate the area slightly. 
We label $l_{ref}$ to denote the image distances calculated on the attribute space of $I_{Ref}$. $l_{ref:recon}$ represents the image distance caused solely by the image reconstruction loss of the generator. $l_{ref:id}$ represents the distance caused by the difference between input identity embeddings of the generator. $l_{ref:recon+id}$ covers both two causes. 
$l_{ref:id}/l_{ref:recon}$ depicts the normalized identity loss.

In the same way, we define: 
\begin{equation}
    \begin{aligned}
    l&_{test:recon} = ||M_{test} \odot (I_{id-test,att-test} - I_{Test})||_2 \\
    l&_{test:recon+id} = ||M_{test} \odot (I_{id-ref,att-test} - I_{Test})||_2 \\
    l&_{test:id} = ||M_{test} \odot (I_{id-test,att-test} - I_{id-ref,att-test})||_2 \\
    \end{aligned}
\end{equation}
where $M_{test}$ is the mask of image $I_{Test}$, $l_{test}$ denotes the image distances calculated on the attribute space of $I_{Test}$.

In addition to the L2 representation of identity loss, we also consider measuring identity loss in angular space since the angular distance is typically a good indicator to evaluate the similarity between identity embeddings.

\begin{figure}
    \centering
    \includegraphics[width=\linewidth]{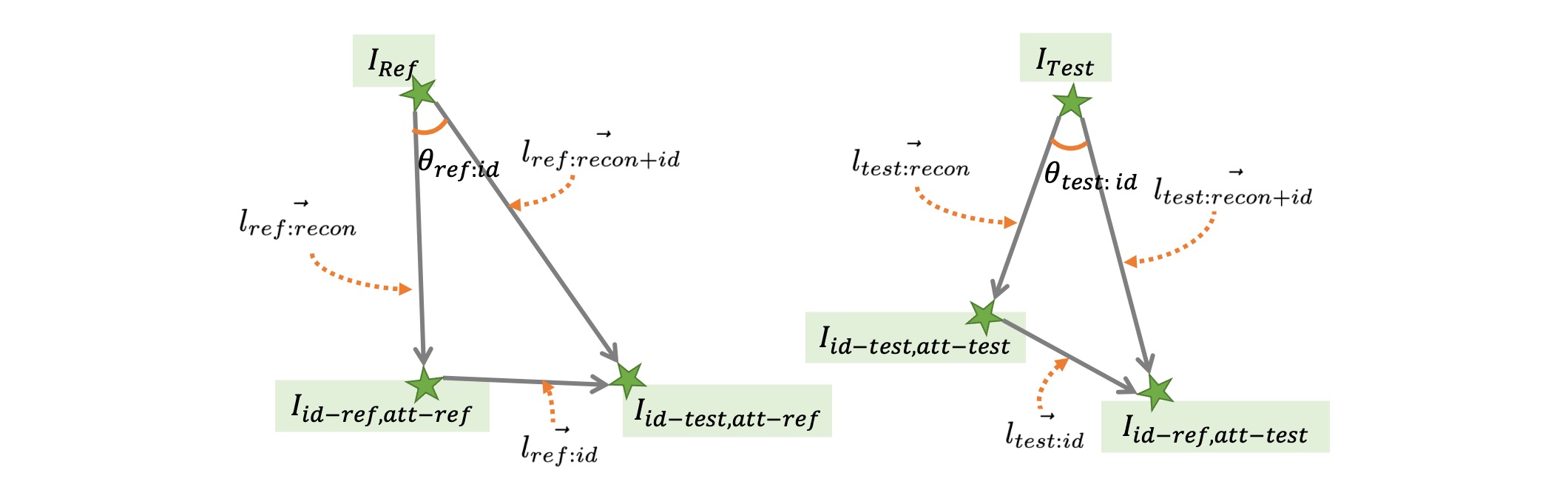}
    \caption{Relationship of aligned generation results and identity loss.}
    \label{fig:vec_relation}
\end{figure}

When vectorizing an image into the vector space, by flattening it from RGB space to a high dimensional vector, the image difference is reformulated as $\vec{l_{ref:recon}} = \vec{M_{ref}} \odot (\vec{I_{id-ref,att-ref}} - \vec{I_{Ref}})$, where $\vec{\cdot}$ means the vector form. 
Accordingly, we have $\vec{l_{ref:recon}}+\vec{l_{ref:id}}=\vec{l_{ref:recon+id}}$. The specific relationship between them is displayed in Figure \ref{fig:vec_relation}. 
We introduce the angle between $\vec{l_{ref:recon}}$ and $\vec{l_{ref:recon+id}}$ as $\theta_{ref:id}$, representing the distance of identity characteristics in angular space. We calculate the value of $\theta_{ref:id}$ following the cosine law.
\begin{equation}
    \theta_{ref:id}=arccos(\frac{{l_{ref:recon}}^2+{l_{ref:recon+id}}^2-{l_{ref:id}}^2}{2\cdot l_{ref:recon}\cdot l_{ref:recon+id}})
\end{equation}
In the same way, we get $\theta_{test:id}$, an identity distance representation from the attribute space of image $I_{Test}$. 

Since the reference and the test image provide different attributes, we fuse the identity difference calculated from four different spaces and therefore design the \textit{Diff-ID} metric to distinguish real and fake images as:
\begin{equation}
    \mathcal{M}=\frac{l_{ref:id}}{l_{ref:recon}}\cdot\frac{l_{test:id}}{l_{test:recon}}\cdot\frac{\theta_{ref:id}+\theta_{test:id}}{2}
\end{equation}

Among them, the normalized losses $l_{ref:id}/l_{ref:recon}$ and $l_{test:id}/l_{test:recon}$ indicate identity differences in Euclidean space. $\theta_{ref:id}$ and $\theta_{test:id}$ represent the differences in the angular space. The identity differences are calculated on the attribute space of both the reference image $I_{Ref}$ and the test image $I_{Test}$. All parts together give a more representative view of identity loss.

\subsection{Adaptive Fine-tuning Scheme for Face-swapping Generator}
\label{section:3.3}
The face-swapping generator is the core function of the image reconstruction module. It transfers the identity of a source face into a target face while keeping the attributes (e.g., expression, posture, lighting, etc.) of the target face unchanged. The
formulation of this process can be written as follows:
\begin{equation}
    I_{result}=G(\mathcal{Z}_{id-source}, \mathcal{Z}_{att-target})
\end{equation}
Off-the-shelf face-swapping generators can be easily deployed into our \textit{Diff-ID} framework.

While the existing generators can function on any two unknown images in the general face-swapping scene, in our \textit{Diff-ID} framework, the generator is applied in the scenario of face changing between two very similar identity characteristics. Therefore, we expect the generator to capture the identity change more sensitively. 
As mentioned in Section~\ref{section:3.2}, the customized metric $\mathcal{M}$ does better in assessing identity loss when the generated result $I_{result}$ meets these two demands: 1) the identity characteristics of $I_{result}$ should be highly consistent with that of the input $I_{source}$; 2) $I_{result}$ should be highly consistent with the input $I_{target}$ in attributes that are unrelated to the identity characteristics, such as facial color consistency, texture, etc.
Therefore, we fine-tune a generator to improve these two consistencies. And we split the fine-tuning task into identity constraint and attribute constraint.

{\bf Identity constraint.} For the identity constraint task, we use an identity preservation loss to preserve the identity of the source image. It is formulated as follows:
\begin{equation}
    \mathcal{L}_{id}=1-\frac{\mathcal{Z}_{id-result}\cdot \mathcal{Z}_{id-source}}{||\mathcal{Z}_{id-result}||_2 ||\mathcal{Z}_{id-source}||_2}
\end{equation}
where $\mathcal{Z}_{id-source}$ represents the identity embedding of the input $I_{source}$, $\mathcal{Z}_{id-result}$ represents the identity embedding of the generation result $I_{result}$.

{\bf Attribute constraint.} For the attribute constraint task, we require the input source face and the target face to have the same identity. In this case, the generated result should look the same as the target face. We calculate the reconstruction loss and use LPIPS\cite{zhang2018unreasonable} for the perceptual loss to capture fine details of images. It is formulated as follows:
\begin{equation}
    \mathcal{L}_{att}=||I_{result}-I_{target}||_1 + \mathcal{L}_{lpips}
\end{equation}

\section{Experiment}
\subsection{Experiment Setting}
\subsubsection{Dataset Setting}
For the task of the generator fine-tuning, we select the first 1,000 images from the CelebA \cite{liu2015deep} as the training set. 
We perform experiments on four well-known face-swapping datasets to evaluate the generalization ability of \textit{Diff-ID}.

{\bf FaceForensics++ (FF++).} FF++ \cite{roessler2019faceforensicspp} is a forensics dataset comprising 1000 real and 4000 fake videos. The fake videos are manipulated with four DeepFake methods: Deepfakes \cite{Deepfakes}, Face2Face \cite{thies2016face2face}, FaceSwap \cite{FaceSwap}, and NeuralTextures \cite{thies2019deferred}. Two compressions are used on the original videos, and two counterparts are generated, resulting in three kinds of video quality: raw, high-quality (HQ), and low-quality (LQ). 
Among them, lower-quality fake videos are more challenging to detect because the forgery traces may be lost in the compression process.
To evaluate the detection of the face-swapping forgeries, we use Deepfakes and FaceSwap to constitute the fake part of the FF++ dataset. Since the other two methods in FF++ belong to the face-reenactment method, which is not within the scope of our approach.

{\bf Google DeepFake Detection (DFD).} DFD \cite{DDD_GoogleJigSaw2019} collects 363 real videos of 28 actors in various scenes and generates over 3000 manipulated videos from them.

{\bf CelebDF.} CelebDF \cite{li2020celeb} includes 590 original videos and 5639 corresponding fake videos of 59 celebrities of different ages, ethnic groups, and genders collected from YouTube.

{\bf DeeperForensics-1.0 (DFo).} DFo \cite{jiang2020deeperforensics} contains source videos collected from 100 paid actors and proposes a DeepFake method to manipulate 1,000 fake videos whose target faces are from the FF++ dataset.

Since \textit{Diff-ID} does not need the DeepFake dataset for training, we do not split the dataset into training or testing but take the whole dataset as the test set. For the image-level evaluation, we sample 20,000 frames from each dataset's real and fake parts. To ensure all the subjects are covered, we sample the same number of images for each identity. For the video-level evaluation, we average the prediction results of the randomly sampled 20 frames as the video detection result. In the following experiments, we report AUC(\%) as the performance metric.

\subsubsection{Implementation Detail} 
We set SimSwap~\cite{chen2020simswap} as the default face-swapping generator. We use MTCNN~\cite{zhang2016joint} to detect the face region in the image. Then, we crop, align, and resize the face at a resolution of the generator's input size. 
To get the face mask, we use the modified BiSeNet\cite{yu2018bisenet} to parse the face region, which is then dilated with the OpenCV \cite{kaehler2016learning} library.

\subsubsection{Reference Image Sampling}
We randomly select one reference image for each identity. Since the FF++ dataset has only one real video per identity, we randomly sample one frame from the real video as the reference image and exclude it from the test set. As for the other three datasets, each identity contains multiple real videos. Therefore, we randomly sample one frame from one real video as the reference image and take the remaining videos as the test set.

\begin{table*}[!t]
\renewcommand{\arraystretch}{1.3} 
\caption{Generalization comparison with SOTA frame-level and video-level methods. Results are reported in AUC scores (\%).}
\label{tab:performance comparison}
\vspace{-2mm}
\centering
\begin{threeparttable}
\begin{tabular}{m{3.2cm}<{\raggedright} m{3.0cm}<{\centering} m{2.2cm}<{\centering} m{2.2cm}<{\centering} m{2.2cm}<{\centering} m{2.2cm}<{\centering}}
\toprule
Method & Training Set & FF++ & DFD & CelebDF & DFo  \\
\hline
\multicolumn{6}{c}{Frame-Level Result}\\
\hline
Xception \cite{chollet2017xception}    & FF++ & {\bf 99.4} &  83.1       & 59.4 & 69.8 \\
DSP-FWA \cite{li2019exposing}      & self collected  & 93.0 & 81.1 & 64.6 & -    \\
Face X-ray \cite{li2020face}   & FF++ (real)$\dag$  & 98.7$\dag$ & \underline{93.5$\dag$} & 74.8 & 72.3 \\
Luo et al. \cite{luo2021generalizing} & FF++  & {\bf 99.4}    & 91.9       & \underline{79.4} & \underline{73.8} \\
SPSL \cite{liu2021spatial}   & FF++  & 98.3  & -   & 76.9 & -    \\
\hline
Diff-ID (ours)  & -  & \underline{99.1}   &  {\bf 98.5}  & {\bf 91.1}  & {\bf 98.3} \\
\hline
\multicolumn{6}{c}{Video-Level Result}\\
\hline
PEL \cite{gu2022exploiting} & WildDeepfake & 61.6$\ddag$ & 86.8 & 82.9 & - \\
DCL \cite{sun2022dual} & FF++ & 99.3 & 91.7 & 82.3 & -\\
PCL+I2G\cite{zhao2021learning} & FF++ (real)$\dag$ &  {\bf 99.9}$\dag$ & {\bf 99.0}$\dag$ & \underline{90.0} & {\bf 99.4} \\
LipForensics\cite{haliassos2021lips} & LRW \& FF++ &  94.9 & - & 82.4 & 97.6 \\
RealForensics\cite{haliassos2022leveraging} & LRW \& FF++   & 98.6 & - & 86.9 & \underline{99.3} \\
\hline
Diff-ID (ours)  & - & \underline{99.5}  & \underline{98.9}  & {\bf 93.1} & 99.2 \\
\bottomrule
\end{tabular}
Data with $\dag$ indicates the model was trained or tested on the raw version of video frames. Data with $\ddag$ indicates the model was trained or tested on the low-quality compressed video frames. Others not specifically marked are tested on high-quality compressed video frames. Top-2 best results are in {\bf bold} and \underline{underlined}.
\end{threeparttable}
\end{table*}

\subsection{Comparisons with Previous Methods}
In this section, the performance of the proposed method is analyzed and compared with other state-of-the-art methods. We test the performance of the proposed method on the four datasets. 

\subsubsection{Comparison on Frame-level Detection}
In this section, we compare \textit{Diff-ID} with the classic and recently proposed methods that focuses on frame-level detection. 
(1) Xception \cite{roessler2019faceforensicspp} is a DeepFake detection method based on XceptionNet model \cite{chollet2017xception}.
(2) DSP-FWA \cite{li2019exposing} uses a CNN model to detect face-warping artifacts introduced by the resizing and interpolation operations in the basic DeepFake manipulation algorithms. This work is trained on self-collected face images.
(3) Face X-ray \cite{li2020face} reveals the blending boundaries in the forged face images. The training data of this algorithm consists of two parts, in which the real images come from the FF++ dataset, and the fake images are self-generated by blending two real images with similar facial landmarks.
(4) Luo et al. \cite{luo2021generalizing} designed functional modules to extract multi-scale high-frequency features and residual guided spatial features to concentrate more on generalizable forgery traces.
(5) SPSL \cite{liu2021spatial} combines spatial image and phase spectrum to capture the up-sampling artifacts of face forgery to improve the transferability of forgery detection.
Methods unspecified are trained on the FF++ dataset of high quality.
The input for the above five methods is only a query face image. For our approach \textit{Diff-ID}, the input is a query image and one reference image of that subject.

As shown in Table \ref{tab:performance comparison}, all methods achieve impressive performance on FF++. It means that the network has the ability to learn discriminative features for known manipulations. However, the performance drops when the network is tested on an unseen manipulation.
For example, Xception trained on FF++ and reports the highest AUC (99.4\%) on FF++. It also behaves well in DFD since the fake samples in DFD are manipulated using Deepfakes, a known forgery method in FF++.
However, when tested on the two other datasets, CelebDF and DFo, Xception faces a catastrophic performance drop (lower than 60\% on CelebDF).
The following four studies, DSP-FWA, Face X-ray, Luo et al., and SPSL, claim to extract common features in forgery images to improve their generalization ability. Compared to Xception, they indeed enhance the detection of unseen manipulations. Luo et al. improved the AUC result by 20\% on CelebDF and 4\% on DFo over Xception. However, all four methods get at most 80\% AUC scores on CelebDF and DFo, showing unsatisfactory cross-dataset generalization ability. 

In contrast, our method achieves good performance on all four datasets. \textit{DIff-ID}'s performance on the FF++ dataset is on par with the best one. Besides, it achieves the best results on the other three datasets, with 98.5\% on DFD, 91.1\% on CelebDF, and 98.3\% on DFo. The satisfactory result of \textit{DIff-ID} mainly benefited from extracting high-level facial identity-inconsistency features as forgery traces and avoiding any facial forgery dataset for training. As a result, \textit{Diff-ID} behaves well on numerous datasets. 
It seems unfair since we obtain additional information from a real image to achieve considerable performance. Nevertheless, getting such a reference image is feasible and conducive for platforms with authenticated users. In a later section, we will compare \textit{Diff-ID} with other approaches using auxiliary reference information to prove its effectiveness further.

\subsubsection{Comparison on Video-level Detection}
In this section, we compare \textit{Diff-ID} with the recently proposed methods that focuses on video-level detection.
(1) PEL \cite{gu2022exploiting} exploits pixel-level and fine-grained frequency-level clues. It uses a progressive enhancement process to facilitate the learning of discriminative face forgery features. We chose the WildDeepfake-trained model for comparison as it has better generalization ability than the one trained on the FF++ (LQ) dataset.
(2) DCL \cite{sun2022dual} performs designed contrastive learning on the constructed positive and negative pairs at different granularities. The method is trained on the FF++ dataset.
(3) PCL+I2G \cite{zhao2021learning} hypothesizes that a forged image contains different source features at different locations. It detects forgery samples by extracting the local source features and measuring their pairwise self-consistency. The algorithm uses an image synthesis approach called inconsistency image generator (I2G) to provide richly annotated training data. Therefore, it needs only real videos from the original videos in FF++ to train the model.
(4) LipForensics \cite{haliassos2021lips} targets high-level semantic irregularities in mouth movements. Through natural lipreading learning, it identifies mouth movements that have undergone facial manipulations as abnormal.
(5) RealForensics \cite{haliassos2022leveraging} exploits the natural correspondence between the visual and auditory modalities in real videos. It learns temporally dense video representations in a self-supervised way and then uses these representations to make real/fake decisions about test videos. 
The above two methods are trained on the FF++ and the LRW dataset \cite{chung2017lip}, which contains 500,000 videos of talking faces with hundreds of different identities.
We also provide video-level results of \textit{Diff-ID} for better comparison with these studies.

The results are shown in Table \ref{tab:performance comparison}. All the methods achieved over 80\% AUC on the CelebDF dataset and over 95\% on the DFo dataset, showing better generalization ability than the frame-level detection methods. These methods use numerous real videos in the training process to learn feature representations, such as facial movements, expressions, and identity, on the continuity and consistency between video frames. As a result, they have achieved good generalization on multiple datasets. 
Among these methods, PCL+I2G and our \textit{Diff-ID} are the two best performers. 
Both of them have reached remarkable results of almost 99\% AUC on the FF++, DFD, and DFo datasets and about 90\% on the CelebDF.
It is worth noting that PCL+I2G is trained and tested on the original uncompressed videos of FF++ and DFD datasets. In comparison, \textit{Diff-ID} gets the score on the compressed high-quality videos, which are much harder to distinguish. 
Even so, \textit{Diff-ID} achieves a very close performance on these two datasets, indicating our excellent detection capabilities. In addition, \textit{Diff-ID} gets the best result on CelebDF, outperforming PCL+I2G by 3.1\% AUC. 
The results demonstrate that although our approach does not utilize the inter-frame information, it still performs better than those that utilize the inter-frame information, confirming that our assessment of identity loss is an effective clue for DeepFake detection.

\subsubsection{Comparison with Reference-assisted Methods}
\begin{table*}[]
    \renewcommand{\arraystretch}{1.3} 
    \caption{Generalization comparison with reference-assisted methods.}
    \label{tab:reference-assisted comparison}
    \vspace{-2mm}
    \resizebox{\linewidth}{!}{
        \begin{tabular}{m{2.4cm}<{\raggedright} m{2.6cm}<{\centering} m{2.8cm}<{\centering} m{1.0cm}<{\centering} m{1.0cm}<{\centering}
        m{1.0cm}<{\centering} m{1.0cm}<{\centering} m{1.6cm}<{\centering} m{1.6cm}<{\centering}}
        \toprule
        \multirow{2}{*}{Method} & \multirow{2}{*}{Training Set} & \multirow{2}{*}{\makecell[c]{Reference\\Per Subject}} & \multicolumn{2}{c}{FF++}  & \multicolumn{2}{c}{DFD} & \multirow{2}{*}{CelebDF} & \multirow{2}{*}{DFo}  \\  \cmidrule(lr){4-5} \cmidrule(lr){6-7}
        
         & & & HQ & LQ & HQ & LQ & & \\ 
        \hline
        \multicolumn{9}{c}{Frame-Level Result}\\
        \hline
        DISC \cite{jiang2021practical}  & FF++/DF &  one image  & 95.7&-  &  98.4 &87.5 & 84.4 & 97.0 \\
        ICT-Ref \cite{dong2022protecting} & MS-Celeb-1M & ten images & 98.6 & - & 93.2 & - & {\bf 94.4} & {\bf 99.3} \\
        ICT-Ref* \cite{dong2022protecting} & MS-Celeb-1M & one image & 98.0 & 96.9 & 92.4 & 87.6 & 85.5 & 98.6 \\
        Diff-ID (ours)   & -  & one image    & {\bf 99.1} & {\bf 97.6}   &  {\bf 98.5}& {\bf 96.1} & 91.1  & 98.3 \\
        \hline
        \multicolumn{9}{c}{Video-Level Result}\\
        \hline
        ID-Reveal\cite{cozzolino2021id} & VoxCeleb2 & videos & 99.0 & {\bf 97.0} & 96.0 &94.0 & 84.0 & - \\
        Diff-ID (ours)  & - & one image & {\bf 99.5} & 96.6  & {\bf 98.9} & {\bf 95.6} & {\bf 93.1} & {\bf 99.2} \\
        \bottomrule
    \end{tabular}
    }
\end{table*}
In this section, we compare our method with reference-assisted studies. We consider two frame-level detection methods, DISC \cite{jiang2021practical} and ICT \cite{dong2022protecting}, and one video-level detection method ID-Reveal \cite{cozzolino2021id}.
(1) DISC constrains the network to focus on the identity-related facial areas, guided by a real reference image, to exploit the intrinsic discriminative forgery clues. The algorithm is trained on Deepfakes, a subset of the FF++ dataset. 
(2) ICT proposes an identity consistency transformer to detect whether a suspect face has identity inconsistency in inner and outer face regions. The network is trained on the MS-Celeb-1M \cite{guo2016ms} dataset that contains 10 million images of 1 million identities. This method can detect forgery either with or without reference images. We will refer to the version with reference images as ICT-Ref in the following.
(3) Based on reference videos of a person, ID-Reveal estimates a temporal biometric embedding of video and uses the embedding to estimate a distance metric to detect fake videos. This approach is trained using the VoxCeleb2 \cite{chung2018voxceleb2} development dataset.

The result in Tabel~\ref{tab:reference-assisted comparison} reveals that all methods show good generalization on different face forgery datasets, owing to the additional information provided by the reference.
DISC provides the detector with an additional real image, which is consistent with our method. In contrast, ICT-Ref randomly samples ten real images for each identity and constructs a reference set. For a fair comparison, we retested the official ICT-Ref model under our reference setting, \textit{one real image for each identity}. The corresponding result is displayed in the line of ICT-Ref*. Under the same reference setting, \textit{Diff-ID} achieves the best performance on various datasets and different video qualities. 
The results of ICT-Ref under different reference settings prove the importance of additional semantic information. For example, comparing the results of the CelebDF dataset, there is a 10\% AUC drop from the ten reference-assisted ICT to the one reference-assisted ICT. The performance drop can be explained by the fact that the videos of each identity contain various scenes of different ages, hairstyles, makeups, or backgrounds, causing diverse representations of inner and outer identity. Therefore, the restricted reference images in ICT-Ref could have limited effects on identity feature consistency comparison. In contrast, \textit{Diff-ID} aligns the images of a specific identity to the same attribute space before identity feature comparison, thus reducing the interference of factors such as hairstyles, makeups, and backgrounds on the identity characteristics.
ID-Reveal is evaluated at the video level using a leave-one-out strategy to constitute the reference dataset. Although with more reference information, its performance on these face-swapping datasets is inferior to our method. Overall, \textit{Diff-ID} shows a preferable performance with the least reference information.

\subsection{Analysis of the method}
\label{section:4.3.1}
In this section, we first compare our method with its naive version ``identity embedding similarity". Then, we discuss the impact of some settings in the method: reference, generator finetuning, mask matrix, and \textit{Diff-ID} metric selection.
\subsubsection{Advantages over Identity Embedding Similarity}
First, we define the specifics of the naive identity embedding similarity approach which is called IESim.

{\bf Face recognition model $\Phi_{id}$.} We apply ArcFace\cite{deng2019arcface} to extract the identity embeddings of face images, which is also the identity extractor of \textit{Diff-ID}.

{\bf Reference selection.} Same as \textit{Diff-ID} that one frame of one real video is randomly sampled as the reference image.

{\bf Method of IESim.} We introduce the identity loss of the test image $I_{Test}$ and the reference image $I_{Ref}$ as:
\begin{equation}
    \mathcal{L}_{id} = 1-cos(\Phi_{id}(I_{Ref}),\Phi_{id}(I_{Test}))
\end{equation}
This identity loss $\mathcal{L}_{id}$ is then used for DeepFake detection. The larger the identity loss is, the more chance the test is a forgery face. 

The comparison results of IESim and \textit{Diff-ID} are reported in Table \ref{tab:IESim}. 
\textit{Diff-ID} performs better than IESim on all three datasets, especially on the DFo dataset. The real images in the DFo dataset are obtained by filming hired actors in short time intervals, excluding the influence of age on the identity characteristics. 
Ideally, it is reasonable that these real images should have the same identity characteristics and be easily distinguished from fake images. However, due to the large variances in poses and expressions in real images, the similarity of identity embeddings between real images could be even more confusing than that with fake images.
In comparison, our method introduces the reconstruction process to align face poses, expressions, etc., to the same attribute space as the reference or the test image. This measure effectively reduces the impact of attribute features that are difficult to disentangle in facial features on identity similarity comparisons. In addition, \textit{Diff-ID} can visualize the face regions where the identity is different by capturing the pixel differences between the aligned generation results, providing more explainable detection results.
    
\begin{table}[!t]
\renewcommand{\arraystretch}{1.3} 
    \caption{Performance comparison of IESim and Diff-ID.}
    \label{tab:IESim}
    \vspace{-2mm}
    \centering
    \begin{tabular}{m{2.0cm}<{\centering} m{1.5cm}<{\centering} m{1.5cm}<{\centering} m{1.5cm}<{\centering}}
        \toprule
         Method & DFD & CelebDF & DFo \\
        \hline
         IESim & 95.9 & 89.8 & 95.0 \\
         Diff-ID & 98.5 & 91.1 & 98.3 \\
         gain & +2.6 & +1.3 & +3.3 \\
         
        \bottomrule
    \end{tabular}
\end{table}

\begin{table}[!t]
    \renewcommand{\arraystretch}{1.3} 
    \caption{Analysis of different reference selection strategies on DFD.}
    \label{tab: ref select strategy}
    \vspace{-2mm}
    \centering
    \begin{tabular}{m{2.5cm}<{\centering} m{3.2cm}<{\centering} m{1.5cm}<{\centering}}
        \toprule
        Select Strategy & AUC & mean \\
        \hline
        random & 98.1 / 97.9 / 98.2 & 98.1 \\
        frontal & {\bf 98.7} / {\bf 98.3} / {\bf 98.8} & {\bf 98.6} \\
        same orientation & 98.2 / 97.9 / 98.5 & 98.2 \\
        \bottomrule
    \end{tabular}
    
\end{table}

\subsubsection{Reference Sensitivity}
Given the diversity of the images in the dataset, it is reasonable to suspect that the selection of reference images will affect the experimental results. In this section, we discuss the impact of different choices of reference images on detection performance.
We mainly consider the influence of head orientation in the reference image, as it will affect facial feature extraction. Specifically, we estimate the orientation of a human head using OpenCV \cite{kaehler2016learning} and Dlib \cite{king2009dlib}. The estimated yaw value can reflect the head orientation.
We evaluate the sensitivity of reference image selection on the DFD dataset, as it contains videos in various poses for each subject. For example, scenes named ``hugging happy" and ``secret conversation" are profile videos. Other scenes, such as ``walking **," show a dynamic process and contain video frames in various poses. Therefore, this dataset meets our needs for face videos with diverse head orientations. 

We randomly select one real video of each identity as the reference pool and then sample 100 real and 100 fake images from the remaining videos to form a test set. Then, we choose one reference image from the reference pool for each test image according to three different yaw selection strategies: 1) The ``random'' strategy, we randomly sample one image from the pool as the baseline; 2) The ``frontal'' strategy, we sample one image with detected yaw between [-5, 5] since a front head orientation usually falls into this range; 3) The ``same orientation'' strategy, we pick one image whose yaw value deviation from the test image is not greater than five. 

Table~\ref{tab: ref select strategy} shows the results of three random experiments.
We find that the ``frontal'' strategy consistently performs best. It is reasonable since more facial features can be revealed and benefit our detection when reconstructing a frontal face. The other two strategies, ``random'' and ``same orientation'', also have a good performance with a drop of no more than 1\% compared to the best result. 
The result proves \textit{Diff-ID} can recognize the identity difference with relatively low sensitivity to the head orientation of reference images. Therefore, in terms of simplicity and practicality, we only sample one random reference image for the whole test sets of each identity in the main experiment.

\subsubsection{Effectiveness of Proposed Gain Strategies} 
In our \textit{Diff-ID}, we design a mask matrix to filter out the noise outside the face region. Besides, we fine-tune the face-swapping generator to meet the needs of identity constraint and attribute constraint. These two strategies in \textit{Diff-ID} aim to improve its capability of capturing identity differences.
In this section, we split each part separately to explore the impact of the two strategies in \textit{Diff-ID}.
Specifically, we conduct the following variants: 1) pre-trained SimSwap as the generator without applying the mask as the baseline. 2) pre-trained SimSwap as the generator with the mask. 3) fine-tuned SimSwap as the generator without the mask. 4) fine-tuned SimSwap as the generator with the mask. 

\begin{table}[!t]
    \renewcommand{\arraystretch}{1.3} 
    \caption{Performance evaluation of proposed gain strategies.}
    \label{tab: component ablation}
    \vspace{-2mm}
    \centering
    \begin{tabular}{m{0.8cm}<{\centering} m{1.4cm}<{\centering} m{0.8cm}<{\centering} m{1.0cm}<{\centering} m{1.0cm}<{\centering} m{1.0cm}<{\centering}}
        \toprule
        Variant&fine-tuning & mask & DFD & CelebDF & DFo\\
        \hline
        1) &  - & - & 98.1 & 90.3 & 97.2\\
        2) & - & $\checkmark $ & 98.3 & 90.2 & 97.6\\
        3) & $\checkmark $ & - &98.3  & 90.5 & 97.5\\
        4) & $\checkmark $ & $\checkmark $ &{\bf 98.5} & {\bf 91.1} & {\bf 98.3} \\
        \bottomrule
    \end{tabular}
\end{table}
The results are reported in Table \ref{tab: component ablation}, and the metric is AUC. 
Comparing variant 1) and variant 3), the adaptive fine-tuning scheme brings about 0.2\% AUC score improvement on average.
Comparing variant 1) and variant 2), as well as variant 3) and variant 4), the performance is further improved by applying the face mask to exclude the noisy attribute loss in the background.
Combining both two components yields the best detection results, with an average performance improvement of 0.8\% over the baseline on the three datasets.

\begin{figure}
    \centering
    \includegraphics[width=\linewidth]{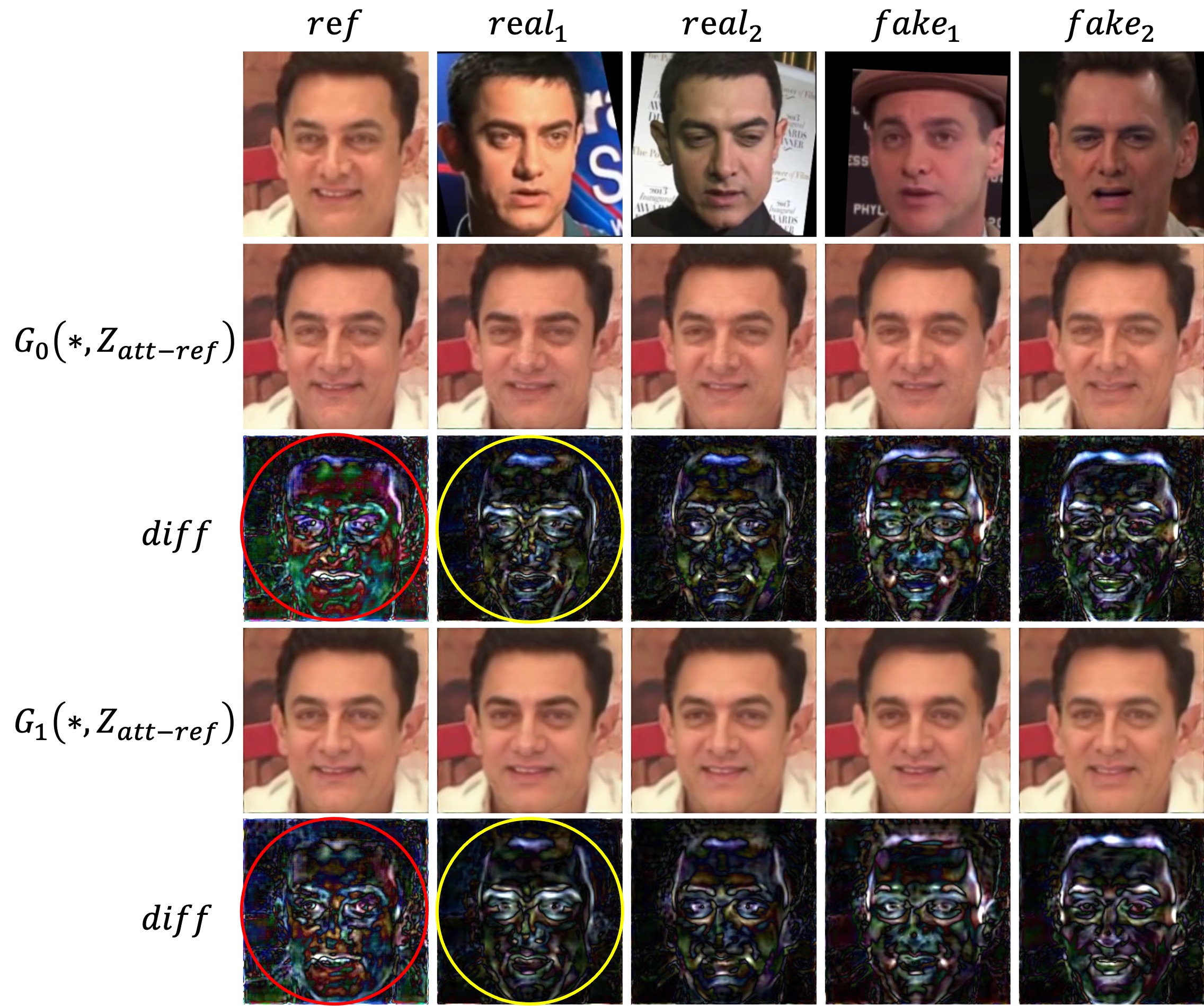}
    \caption{Face-swapping results comparison of the pre-trained and the fine-tuned generator. The diff images are magnified ten times for better illustration.}
    \label{fig:recdiff}
\end{figure}
In order to better explain the performance improvement brought by our fine-tuning strategy, in Figure~\ref{fig:recdiff}, we show the comparison between the pre-trained model (represented as $G_{0}$) and our fine-tuned model (represented as $G_{1}$) in terms of attribute feature preservation and identity difference extraction.
The first row displays face images from a specific identity, such as Aamir Khan. In detail, there is one reference image, two real images, and two fake images in sequence.
The second row shows the face-swapping results where the reference provides the attribute embedding, the five examples in the first row provide the identity embeddings, and the pre-trained SimSwap $G_{0}$ serves as the generator.
The third-row results are the image pixel differences. Specifically, the first one is $diff(I_{id-ref,att-ref}, I_{ref})$, indicating the reconstruction loss of the pre-trained generator $G_{0}$. The following four are $diff(I_{id-test,att-ref}, I_{id-ref,att-ref})$, where $test$ refers to the first-row images.
The brighter areas in the $diff$ images disclose the noticeable attribute or identity loss.
We expect that the attribute loss obtained in the $diff$ image is small enough while the identity loss is large enough, such that we can more accurately assess the identity difference between the original reference image and the test image.
The images in the last two rows are generated in the same way but use our fine-tuned SimSwap $G_{1}$ as the generator. Comparing the two red-circled $diff$ images, we find that the pixel values in the lower image are smaller, especially in the identity-insensitive regions such as the forehead, cheeks, and facial contours. It proves the fine-tuned generator better maintains attribute features in the generation results.
Meanwhile, the fine-tuned generator maintains the ability to reveal differences in input identity features as the pre-trained generator. This can be seen by comparing the two $diff$ images circled in yellow since there is little difference in identity-sensitive areas. Similar observations can be found on that of the other real image ($real_2$).
Moreover, our fine-tuned generator more clearly and explicitly reveals that on which facial characteristics the fake image differs from the reference image on the $diff$ images. For example, the eyes and mouth corners of $fake_1$ appear different from the reference image; the eyebrows of $fake_2$ appear different. Due to the better suppression of the noise caused by the attribute loss, the identity difference is better recognized in the $diff$ images.
Overall, the fine-tuned generator performs better on identity and attribute constraints. It is easier to capture minor identity differences in the input and reflect them in the generated images. Therefore, the fine-tuned generator with stricter constraints on identity and attributes plays a great role in the task of \textit{Diff-ID} to detect fake images.

\subsubsection{Customized Metric Evaluation}
In this section, we conduct experiments to verify the effectiveness of different metrics in distinguishing real and fake images. 
As mentioned before, we calculate $l_{ref:id}/l_{ref:recon}$ and $l_{test:id}/l_{test:recon}$ to normalize the identity differences between the reference and the test image. We also use angle $\theta_{ref:id}$ and $\theta_{test:id}$ to represent the distance of the two identity features in the angular space. 
The result in Table \ref{tab: metric ablation} indicates that these metrics behave well and are robust among diverse datasets. 

\begin{table}[!t]
    \renewcommand{\arraystretch}{1.3} 
    \caption{Effectiveness of different quantitative metrics.}
    \label{tab: metric ablation}
    \vspace{-2mm}
    \centering
    \begin{tabular}{m{0.4cm}<{\centering} m{0.8cm}<{\centering} m{0.5cm}<{\centering} m{0.9cm}<{\centering} m{0.5cm}<{\centering} m{0.7cm}<{\centering} m{0.7cm}<{\centering} m{0.7cm}<{\centering}}
        \toprule
        {\scriptsize $l_{ref:id}$} & {\scriptsize $\frac{l_{ref:id}}{l_{ref:recon}}$} & {\scriptsize $\theta_{ref:id}$} & {\scriptsize $\frac{l_{test:id}}{l_{test:recon}}$} & {\scriptsize $\theta_{test:id}$} & {\footnotesize DFD} & {\footnotesize CelebDF} & {\footnotesize DFo}\\
        \hline
        $\checkmark$ & - & - & - & - & 93.9 &83.5 & 86.1 \\
        - & $\checkmark$ & - & - & - &97.6 & 88.2&  96.6\\
        - & - &  $\checkmark$ & - & - &97.7 &86.7 &  96.0\\
        - & $\checkmark$ & $\checkmark$ & - & - &97.7 & 88.1 & 96.5\\
        - & $\checkmark$ & - &$\checkmark$ & - & 98.6& 91.2 & 98.4\\
        - & $\checkmark$ &$\checkmark$  &$\checkmark$ & $\checkmark$ & 98.5& 91.1 & 98.3\\
        \bottomrule
    \end{tabular}
\end{table}

The result reveals that the performance of metric $l_{ref:id}$ is not as good as $l_{ref:id}/l_{ref:recon}$. 
It proves that an isolated difference does not represent the identity characteristics loss well, while the normalized one does. In the following, we give a detailed explanation of this result. $l_{ref:id}$ (so does $l_{test:id}$) represents the difference between the outputs generated from two similar identity features and assesses the identity differences between the reference and test images. However, the value of $l_{ref:id}$ is largely related to the quality of the reference image besides the ability of the face-swapping generator. With the generator fixed, fine-grained attributes of high-quality reference images disclose more identity differences, resulting in a larger $l_{ref:id}$ compared with the low-quality ones. On the other hand, delicate image details might exceed the generator's reconstruction ability and be reflected in a larger reconstruction loss $l_{ref:recon}$.
Therefore, the normalized metric $l_{ref:id}/l_{ref:recon}$ gives a fair comparison when the quality of reference images varies, and the result confirms it. 

The $\theta_{ref:id}$ performs as well as the normalized loss $l_{ref:id}/l_{ref:recon}$. They could complement each other, as one indicates an identity difference in numerical value and the other in angular space. Thus, our \textit{Diff-ID} metric merges the identity loss from two different dimensions and is well-behaved in DeepFake detection.

\subsection{Generalization to Other Face-swapping Models.}
In this section, we analyze the generalization ability of \textit{Diff-ID} with other face-swapping generators.
We replace the backbone generator with FaceShifter \cite{Li2020faseshifter} or HifiFace \cite{ijcai2021-157} because of the following considerations.
\begin{itemize}[leftmargin=*]
\item First, FaceShifter and HifiFace achieve good ID retrieval scores, meaning that the generated results are highly similar to the source in terms of identity features. Therefore, they are more likely to satisfy \textit{Diff-ID}'s demand of identifying very subtle identity differences. If a great identity characteristics change occurs in face swapping, it will damage the assessment of identity loss.
\item Second, FaceShifter and HifiFace adopt two typical identity feature extractors, respectively. In specific, FaceShifter uses the face recognition model as an identity encoder to extract the identity embeddings from a 2D face directly. Following another direction of work, HifiFace uses a 3D face reconstruction network to rebuild 2D faces into 3D faces and obtains identity-related representation through the 3D Morphable Models (3DMM).
\item Third, FaceShifter and HifiFace are relatively computation friendly. Other recent methods, such as MegaFS and HiRes, use StyleGAN as a Decoder to generate high-resolution (1024*1024) images. However, the faces in the DeepFake dataset are not that high-resolution. In this case, FaceShifter and HifiFace can better balance the image quality and computation overhead.
\end{itemize}  

Regarding specific model deployment details, the FaceShifter model is an unofficial implementation of our own. According to the FaceShifter paper, the two networks for the entire pipeline are AEI-Net and HEAR-Net. We only implement AEI-Net, the face-swapping network that plays the main role. The HifiFace model we use is an unofficial implementation of MINDsLab \cite{HifiFace}. Since HifiFace predicts the face mask itself and confines the face modification region inside the mask, we only apply the fine-tuning strategy to it. Furthermore, we adjust our image processing to fit the resolution of FaceShifter and HifiFace (256$\times$256).

We conduct experiments on CelebDF since it is the currently most challenging DeepFake dataset for \textit{Diff-ID} or other state-of-the-art methods. In CelebDF, each identity owns both real videos with various backgrounds and rich fake videos, similar to real-world forgery detection scenarios. Therefore, if \textit{Diff-ID} with an appropriate generator can perform well on the CelebDF dataset, then it will also have good detection ability in practical applications.

\begin{table}[!t]
    \renewcommand{\arraystretch}{1.3} 
    \caption{\textit{Diff-ID}'s performance with other backbone generators.}
    \label{tab:generalize to other generators}
    \vspace{-2mm}
    \centering
    \begin{tabular}{m{1.8cm}<{\centering} m{1.2cm}<{\centering} | m{1.8cm}<{\centering} m{0.8cm}<{\centering} m{1.2cm}<{\centering}}
        \toprule
        \multicolumn{2}{c|}{HifiFace} & \multicolumn{3}{c}{FaceShifter} \\ \hline
        fine-tuning & CelebDF &fine-tuning & mask & CelebDF\\ \hline
        \multirow{2}{*}{-}& \multirow{2}{*}{79.5}&-&-&86.6\\
         & & - & $\checkmark$ & 85.6\\
        \multirow{2}{*}{$\checkmark$}&\multirow{2}{*}{82.7} & $\checkmark$ & - & 86.8 \\
          & & $\checkmark$ & $\checkmark$ & 87.2 \\
        \bottomrule
    \end{tabular}
\end{table}
The evaluation results are shown in Table~\ref{tab:generalize to other generators}. \textit{Diff-ID} with both FaceShifter and HifiFace achieves high AUC scores above 80\%, proving that it generalizes easily to these two generators. 
It is noted that the pre-trained FaceShifter with mask faces a slight performance drop. We speculate that this is because the incomplete version of FaceShifter does not preserve the attributes well enough. When fake images have a different facial contour (e.g., hairline) from the real, it may be added to the identity loss but excluded by the mask. For example, fusing a larger fake forehead into the reference may compromise the preservation of hairstyle attributes, for a conflict of identity (large forehead) and attributes occurs. As a result, the differences in hair generation are accumulated into the identity loss but are excluded if there is a reference face mask. 
On the contrary, the fine-tuned FaceShifter reduces attribute loss (e.g., hairline differences) and concentrates more on the identity difference in the central face. In this circumstance, the mask matrix filters out the noise outside face regions without affecting identity-induced image differences. Therefore, adaptive fine-tuning together with the mask achieves the best performance.

\begin{figure}
    \centering
    \includegraphics[width=\linewidth]{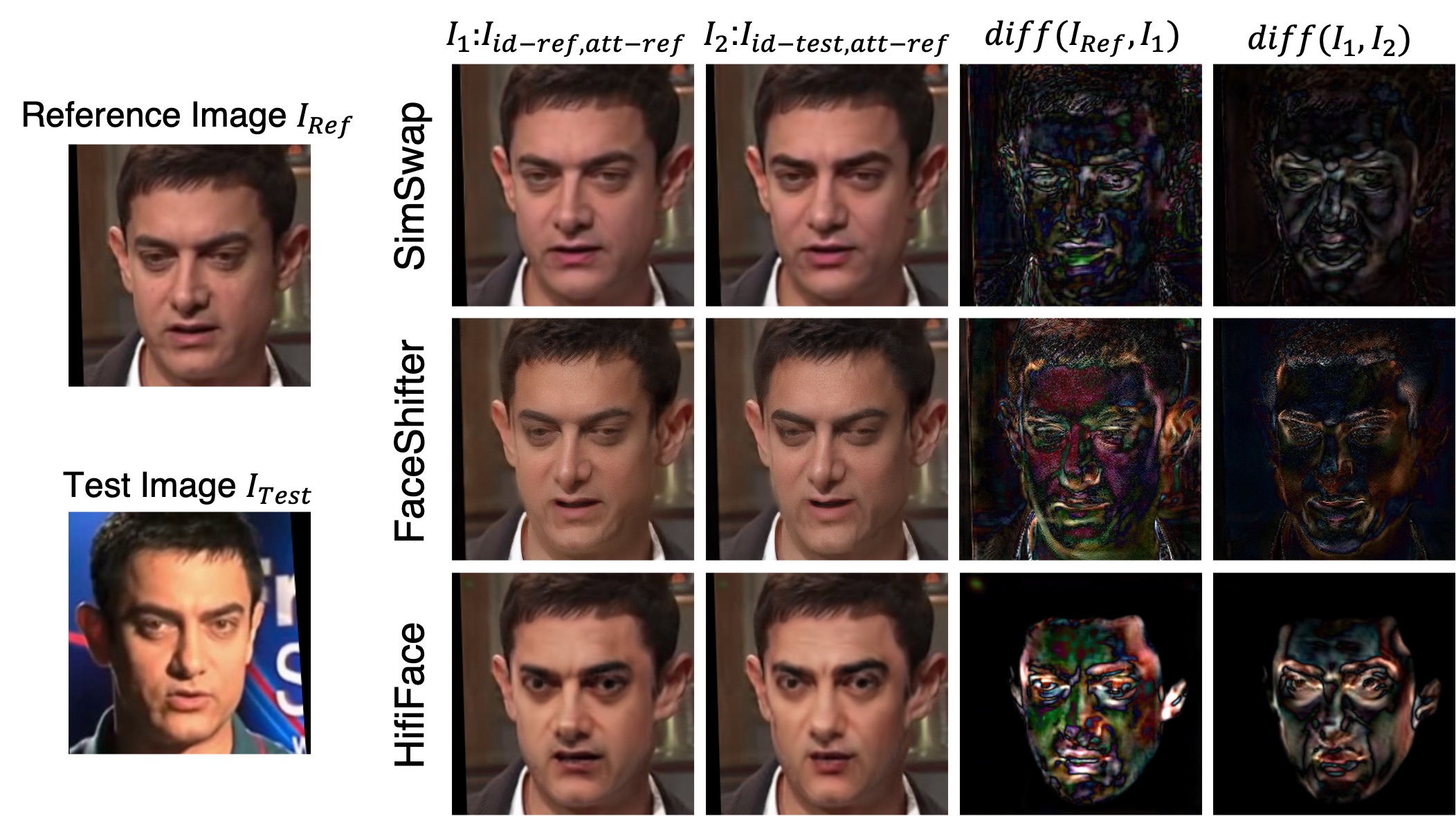}
    \caption{Face-swapping results of different generators.}
    \label{fig:diffGAN}
\end{figure}
In general, \textit{Diff-ID} can be compatible with different face-swapping generators if the generator can preserve the identity and attribute features of the input in the generated results. With different generators as backbones, \textit{Diff-ID} behaves slightly differently.
Figure \ref{fig:diffGAN} displays the face-swapping results of SimSwap, FaceShifter, and HifiFace. From the images, we can find that SimSwap best preserves the input attribute features in the generated results ($I_1$ and $I_2$), followed by FaceShifter, and HifiFace is the worst.
Specifically, the image $I_1$ is the self-reconstruction of the reference image $I_{Ref}$, and the image $diff(I_{Ref}, I_1)$ mainly reflects the ability of the generator to maintain the attribute features. Brighter $diff$ images represent the greater loss of attributes, which will degrade the performance of \textit{Diff-ID}.
Furthermore, we find that SimSwap best captures the identity differences between reference and test images, which are shown in $diff(I_1, I_2)$. 
Compared with FaceShifter, the $diff$ result of SimSwap is brighter and more concentrated in identity-sensitive face regions, indicating a better ability to capture the identity differences. 
Although the $diff$ result of HifiFace is the brightest, it also has large values in identity-insensitive regions such as the forehead and cheeks, indicating that identity loss is polluted by attribute loss. In this case, \textit{Diff-ID} cannot accurately assess identity differences, and its ability to detect fake faces is thus weakened.
Based on the above reasons, SimSwap outperforms FaceShifter and HifiFace.
In conclusion, \textit{Diff-ID} can do better in DeepFake detection if it integrates a generator that can preserve attribute features well and is sensitive to identity features.

\subsection{Method Robustness to Image Quality Distortion.}
When faced with image distortions such as compression, previous works suffer from a drastic drop in their detection performance. Cozzolino et al. \cite{cozzolino2021id} mentioned in their study that ``it is possible to observe a sharp performance degradation of most methods in the presence of strong compression." Other studies like LipForensics \cite{haliassos2021lips} and ADD \cite{le2021add} have also highlighted the problem that DeepFake detection methods are adversely affected by compression. 
As the image quality decreases, forgery details may be lost, and thus the forgery information obtained by the detection network will be weakened or even eliminated. 

\begin{figure}
    \centering
    \includegraphics[width=\linewidth]{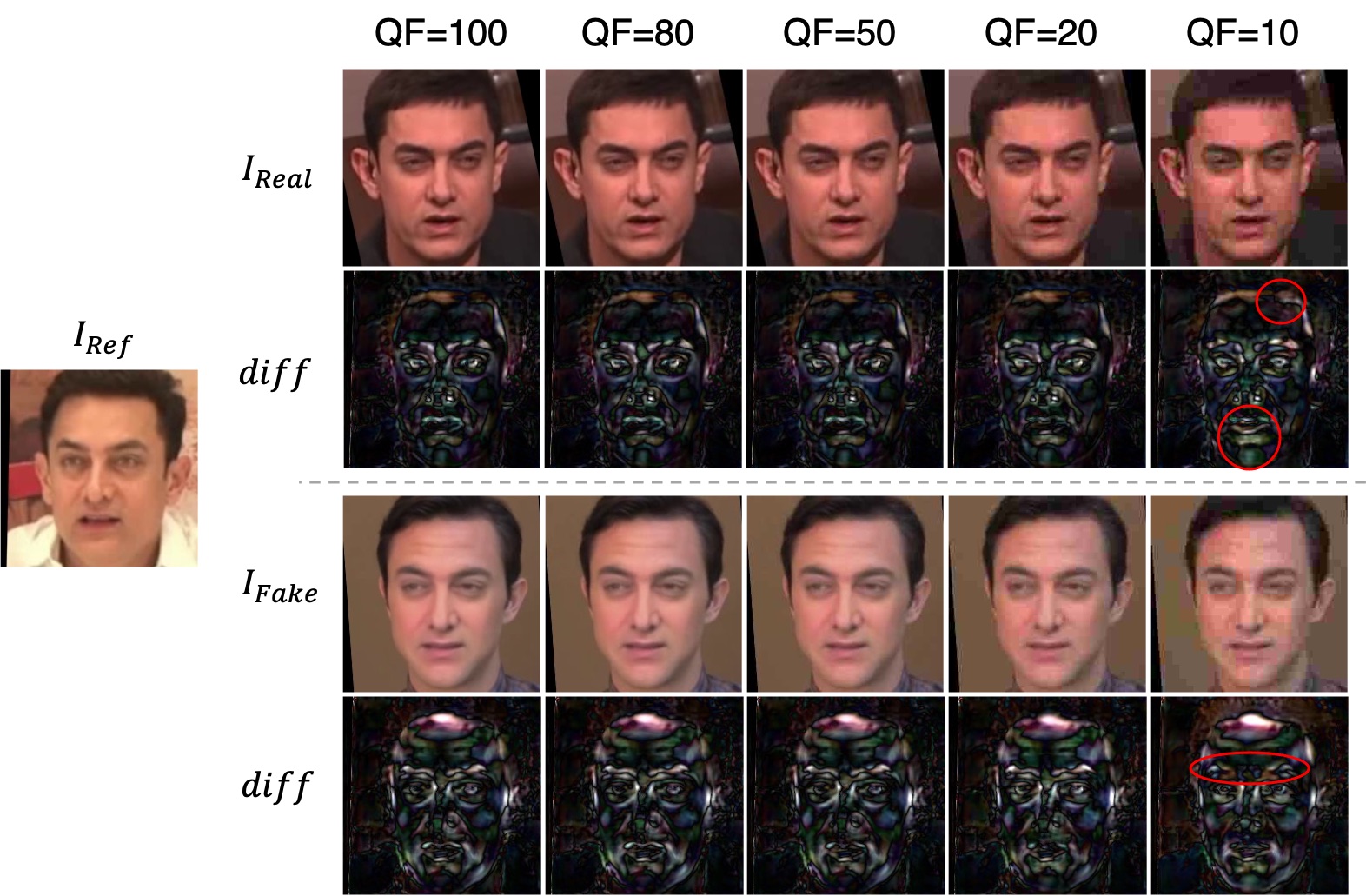}
    \caption{Example results under five different JPEG Quality Factors (QF). The diff images are magnified ten times for better illustration. The noticeable changes mentioned in the main text are circled.}
    \label{fig:metric_QF}
\end{figure}
In this section, we evaluate the stability of \textit{Diff-ID} when the quality of the test image degrades. The JPEG compression quality is usually summarized using the JPEG Quality Factor (QF), which has a value from 0 to 100. 
Figure~\ref{fig:metric_QF} illustrates the identity differences captured from the face images of Aamir Khan at different compression qualities.
On the far left is the reference image used for identity difference comparison. The test images get more compressed from left to right, where QF=100 means uncompression, and QF=10 means heavy compression. The resulting image differences $diff(I_{id-test,att-ref}, I_{id-ref,att-ref})$ following the \textit{Diff-ID} workflow are shown below the compressed test images.
As a high-level semantic feature, we find that identity shows its invariance under image compression. With compression, some details in the image may be lost, while identity features remain almost unchanged. In the figure, when the test image is compressed from QF=100 to QF=20, a pronounced grid-like texture appears, and the colors of the adjacent areas become no longer smooth, which leads to the loss of many detailed features. However, the differences in identity features captured on the $diff$ image still point to the same face region. There is little difference between the $diff$ image with QF=20 and the image with QF=100. Only when the image is catastrophically compressed to QF=10 does the $diff$ image show a noticeable change identified by the red circles. In this case, the images are almost mosaic, with severely distorted facial contours, indicating a large loss of identity details.

To more fine-grained evaluate the robustness of the \textit{Diff-ID} metric to image compression, we randomly sample 200 real and 200 fake images of one person from the CelebDF dataset. For each test image $I_{T}$, we compress it with different JPEG qualities and get a set of variants $\{I_{T,QF=20}, I_{T,QF=25}, ..., I_{T,QF=95}, I_{T,QF=100}\}$. Then, we compute their \textit{Diff-ID} metric and denote the results as $\{ \mathcal{M}_{QF=20},\mathcal{M}_{QF=25}, ..., \mathcal{M}_{QF=95}, \mathcal{M}_{QF=100} \}$. We ignore compression below QF=20 because the image quality hardly degrades to this level, even after multiple transmissions over the Internet.
\begin{figure}
    \centering
    \includegraphics[width=\linewidth]{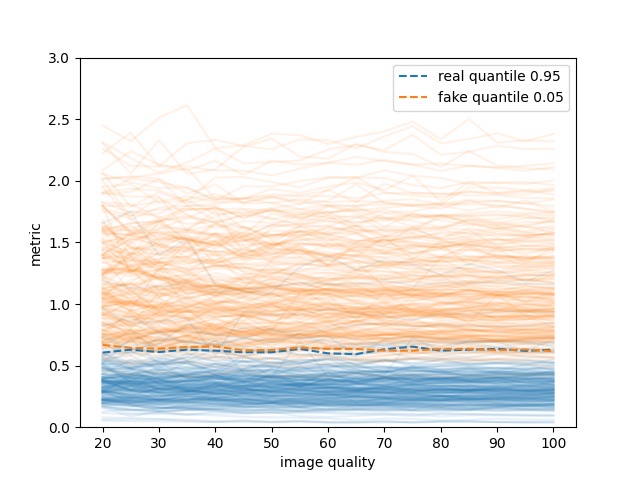}
    \caption{\textit{Diff-ID} metric change curve with respect to JPEG quality factor of 400 test images. Real in blue, Fake in orange. The ninety-fifth quantile change curve of the real images and the fifth quantile change curve of the fake images are marked in the figure.}
    \label{fig:metric_eva}   
\end{figure}
As shown in Figure~\ref{fig:metric_eva}, we plot the \textit{Diff-ID} metric change curves of real and fake images in blue and orange, respectively, where each polyline represents a set of metrics of a specific test image. In addition, we marked the 95\% quantile change curve of the real image and the 5\% quantile change curve of the fake image in the figure. According to the illustrated results, we can conclude that no matter how the image is compressed, most of the test images can be accurately distinguished as real or fake by a threshold of about 0.6. It demonstrates that our \textit{Diff-ID} metric is robust to image compression. 
\begin{table}[!t]
    \renewcommand{\arraystretch}{1.3} 
    \caption{Robust analysis with respect to JPEG quality factor.}
    \label{tab: compression auc}
    \vspace{-2mm}
    \centering
    \begin{tabular}{m{1.2cm}<{\centering} m{0.7cm}<{\centering}
    m{0.7cm}<{\centering}
    m{0.7cm}<{\centering} m{0.7cm}<{\centering} m{0.7cm}<{\centering} m{1.2cm}<{\centering}}
        \toprule
        \multirow{2}{*}{Dataset} & \multicolumn{5}{c}{Quality Factor} & \multirow{2}{*}{$\Delta$AUC}\\
        \cmidrule{2-6}
         & 100 & 80 & 60 & 40 & 20 & \\
        \hline
        DFD & 98.5 & 98.3 & 98.2 &98.1 & 97.9 & -0.6\\
        CelebDF & 91.1 & 90.7 & 90.3 & 90.0 & 89.1 & -2.0 \\
        DFo & 98.3 & 98.1 & 97.8 & 97.5 & 97.1 & -1.2\\
        \bottomrule
    \end{tabular}
\end{table}

We further conduct experiments on the full CelebDF dataset and two other datasets, DFD and DFo. Table \ref{tab: compression auc} reports the detection performance of \textit{Diff-ID} when different compressions are applied to the test images. The results show that the performance drop due to image compression is no more than 2\%AUC, confirming the robustness of our method.
In summary, the robustness of the \textit{Diff-ID} metric to image distortions benefits from several aspects. On the one hand, \textit{Diff-ID} incorporates a well-trained face-swapping generator to stabilize the quality of the reconstructed images. On the other hand, \textit{Diff-ID} mines high-level semantic identity features extracted stably under image compression.

\section{Discussion and Limitation}
{\bf Other Face Forgery Types.} Our detection method is designed for face swapping, one of the various types of facial forgery. We developed our approach based on the finding that swapped fake faces suffer identity loss compared to real faces. Due to the entanglement of identity and attribute on the facial characteristics, the identity feature of the source face in the face-swapping process inevitably conflicts with the target attribute feature, resulting in a non-negligible identity loss for the swapped fake faces. However, other forgery types, such as face reenactment, where only expressions and mouth shapes are changed, result in negligible identity loss. Our approach struggles to capture such subtle identity changes. Considering the diversity of face forgery algorithms, combining multiple methods to detect forged samples in practical applications is expected. Our method performs well on face-swapping detection and can be used together with other methods to tackle the problem of DeepFake detection.

{\bf Dependant on the Underlying Face-swapping Generator.} Our method seems to rely heavily on the capabilities of the underlying face-swapping generator. However, \textit{Diff-ID} is not limited to a specific generator but can be easily deployed with different generators. As illustrated in the experimental section, the generator's sensitivity to identity features, as well as its ability to preserve identity and attribute features across face-swapping results, affect \textit{Diff-ID}'s ability to capture identity loss. Therefore, we design the fine-tuning strategy for the generator to improve its capabilities. Modeling a generator suitable for swapping faces of similar identities can further enhance the performance of \textit{Diff-ID}, which will be our future work.

\section{Conclusion}
In this work, we provide insights into face-swapping detection in that forgery images exhibit an inevitable identity loss from the source real image caused by feature fusion conflicts. Based on this, we propose an explainable forgery detection scheme. Specifically, we introduce \textit{Diff-ID}, which incorporates a face-swapping generator to reconstruct aligned image pairs that visualize identity differences between a test image and an authentic reference image. Then, we design some complementary components that are beneficial for capturing identity differences and reducing noisy attribute loss. Finally, identity inconsistency is quantified via a customized metric to distinguish fake images from real ones. Extensive experiments prove that our method has achieved good detection results on multiple datasets compared to other SOTA approaches. Meanwhile, it has better generalization ability for unknown forgery algorithms and prominent robustness to image compression.

\ifCLASSOPTIONcaptionsoff
  \newpage
\fi



%



\bibliographystyle{IEEEtran}
\bibliography{mybibfile}

%








\end{document}